\documentclass[sigconf]{aamas}  

\usepackage{booktabs}

\setcopyright{ifaamas}  
\acmDOI{doi}  
\acmISBN{}  
\acmConference[AAMAS'18]{Proc.\@ of the 17th International Conference on Autonomous Agents and Multiagent Systems (AAMAS 2018), M.~Dastani, G.~Sukthankar, E.~Andre, S.~Koenig (eds.)}{July 2018}{Stockholm, Sweden}  
\acmYear{2018}  
\copyrightyear{2018}  
\acmPrice{}  

\usepackage{gensymb}
\usepackage{booktabs}
\usepackage{amssymb,verbatim,epstopdf,algorithm,algcompatible,amsmath,caption,subcaption,algcompatible}
\usepackage[font=small,skip=0pt]{caption}

\renewcommand\footnotetextcopyrightpermission[1]{} 
\begin{document}
\title{Self-Organizing Maps as a Storage and Transfer Mechanism in Reinforcement Learning}  



\author{Thommen George Karimpanal}
\affiliation{Singapore University of Technology and Design}
\email{thommen_george@mymail.sutd.edu.sg}
\author{Roland Bouffanais}  
\affiliation{Singapore University of Technology and Design}
\email{bouffanais@sutd.edu.sg}

\begin{abstract}
The idea of reusing information from previously learned tasks (source tasks) for the learning of new tasks (target tasks) has the potential to significantly improve the sample efficiency reinforcement learning agents. In this work, we describe an approach to concisely store and represent learned task knowledge, and reuse it by allowing it to guide the exploration of an agent while it learns new tasks. In order to do so, we use a measure of similarity that is defined directly in the space of parameterized representations of the value functions. 
This similarity measure is also used as a basis for a variant of the \textit{growing self-organizing map} algorithm, which is simultaneously used to enable the storage of previously acquired task knowledge in an adaptive and scalable manner. We empirically validate our approach in a simulated navigation environment 
and discuss possible extensions to this approach along with potential applications where it could be particularly useful.
\end{abstract}

%


\maketitle

\section{Introduction}

The use of off-policy algorithms~\cite{geist2014off} in reinforcement learning (RL)~\cite{sutton2011reinforcement} has enabled the learning of multiple tasks in parallel. This is particularly useful for agents operating in the real-world, where a number of tasks are likely to be encountered, and may be required to be learned~\cite{sutton2011horde,white2012scaling,KARIMPANAL201739}. Ideally, as an agent learns more and more tasks through its interactions with the environment, it should be able to efficiently store and extract meaningful information, which could be useful for accelerating its learning on new, possibly related tasks. 
This area of research, which aims at addressing the issue of effectively reusing previously accumulated knowledge is referred to as transfer learning \cite{taylor_transfer_2009}. 

Formally, transfer learning is an approach to improve learning performance on a new `target' task $M_{T}$, using accumulated knowledge from a set of `source' tasks, $M_{S}=\{M_{s_{1}}...M_{s_{i}}...M_{s_{n}}\}$. Here, each task $M$ is a \textit{Markov Decision Process (MDP)}~\cite{Puterman:1994:MDP:528623}, such that $M=\{\mathcal S,\mathcal A,\mathcal T,\mathcal R\}$, where $\mathcal S$ is the state space, $\mathcal A$ is the action space, $\mathcal T$ is the transition function, and $\mathcal R$ is the reward function. In this work, we address the relatively simple case where tasks vary only in the reward function $\mathcal R$, while $\mathcal S, \mathcal A$ and $\mathcal T$ remain fixed across the tasks. 
For knowledge transfer to be effective, source tasks need to be selected appropriately. Reusing knowledge from an inappropriately selected source task could lead to negative transfer \cite{Lazaric2012,taylor_transfer_2009}, which is detrimental to the learning of the target task. In order to avoid such problems and ensure beneficial knowledge transfer, a number of MDP similarity metrics \cite{ferns2004metrics,carroll2005task} have been proposed.  
However, it has been shown that the utility of a particular MDP similarity metric depends on the type of transfer mechanism used ~\cite{carroll2005task}. In addition, these transfer mechanisms are generally not designed to handle situations involving a large number of source tasks. This could be limiting for both embodied as well as virtual agents operating in the real-world.
For such an agent, the value functions pertaining to hundreds or thousands of tasks may be learned over a period of time. Some of these tasks may be very similar to each other, which could result in considerable redundancy in the stored value function information. From a continual learning perspective, a suitable mechanism may be needed to enable the storage of such information in a scalable manner. In the approach described here, the knowledge of a task is assumed to be contained in the value function ($Q$-function) associated with it. We assume that these value functions are represented using parameter weights, which are learned from the agent's interactions with its environment. We define a cosine similarity metric within this value function weight (parameter) space, and use this as a basis for maintaining a scalable knowledge base, while simultaneously using it to perform knowledge transfer across tasks. 

The proposed mechanism enables the storage of value function weight vectors using a variant of the growing self organizing map (GSOM)\cite{alahakoon2000dynamic}. The inputs to this GSOM algorithm consist of
the value function weights of new tasks, along with any representative value function weights extracted from previously learned tasks. The resulting map would ideally correspond to value function weights representative of previously acquired task knowledge, topologically arranged in accordance with their relation to each other. 
As the agent interacts with its environment and learns the value function weights corresponding to new tasks, this new information is incorporated into the SOM, which 
evolves by growing to a suitable size in order to sufficiently represent all of the agent's gathered knowledge. 
Each element/node of the resulting map is a variant of the input value function weights (knowledge of previously learned tasks). These variants are treated as solutions to arbitrary source tasks, each of which is related to some degree to one of the previously learned tasks.  The aim of storing knowledge in this manner is not to retain the exact value function information corresponding to all the previously learned tasks, but to maintain a compressed and scalable knowledge base that can approximate the value function weights of the previously learned tasks. 

While learning a new target task, this knowledge base is used to identify the most relevant source task, based on the same similarity metric. 
The value function associated with this task is then greedily exploited to provide the agent with action advice to 
guide it towards achieving the target task. Due to random initialization, the agent's initial estimates of the value function weights corresponding to the target task is poor. However, as it gathers more experience through its interactions with the environment, these estimates improve, which consequently improves its estimates of the similarities between the target and source tasks. As a result, the agent becomes more likely to receive relevant action advice from a closely related source task. This action advice can be adopted, for instance, on an $\epsilon$-greedy basis, essentially substituting
the agent's exploration strategy. In this way, the knowledge of source tasks is used to merely guide the agent's exploratory behavior, thereby minimizing the risk of negative transfer which could have otherwise occurred especially if value functions or representations were directly transferred between the tasks. 

\begin{figure}[ht]
\centering
\includegraphics[width=0.8\linewidth]{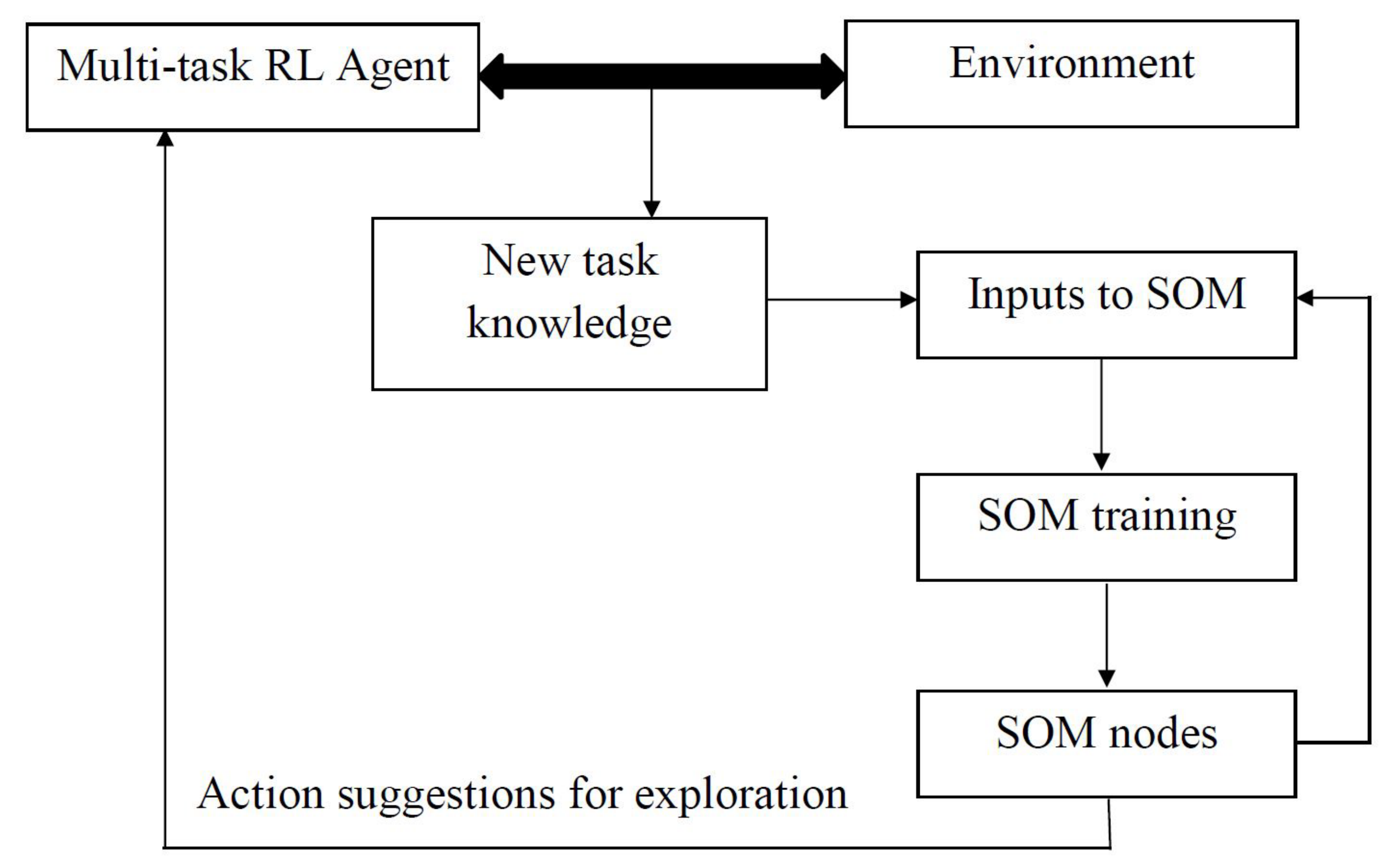}
\caption{The overall structure of the proposed approach} 
\label{gen_struct}
\end{figure}

Apart from maintaining an adaptive knowledge base of value function weights related to previously learned tasks, the proposed approach aims to leverage this knowledge base to make informed exploration decisions, which could lead to faster learning of target tasks. This could be especially useful in real-world scenarios where factors such as learning speed and sample efficiency are critical, and where several new tasks may need to be learned continuously, as and when they are encountered. The overall structure of the proposed methodology is depicted in Figure \ref{gen_struct}.

\section{Related Work}

The sample efficiency of RL algorithms is one of the most critical aspects that determines the feasibility of its deployment in real-world applications. Transfer learning is one of the mechanisms through which this can be addressed. Consequently, numerous techniques have been proposed \cite{Lazaric2012,taylor_transfer_2009,zhan2015online} to efficiently reuse the knowledge of learned tasks. A number of these \cite{carroll2005task,ammar2014automated,song2016measuring} rely on a measure of similarity between MDPs in order to choose an appropriate source task to transfer from. However, this can be problematic, as no such universal metric exists \cite{carroll2005task}, and some of the useful ones may be computationally expensive \cite{ammar2014automated}. Here, the similarity metric used is computationally inexpensive, and the degree of similarity between two tasks is based solely on the value function weights associated with them. Also, in the approach described here, once an appropriate source task is identified, its value functions are used solely to extract action advice, which is used to guide the exploration of the agent. Similar approaches to transfer learning using action advice exist \cite{torrey2013teaching,zhan2015online,zimmer2014teacher}, where a teacher-student framework for RL is adopted. 
The transfer mechanism described here is similar in principle, but is inherently tied to the SOM-based approach for maintaining the knowledge of learned tasks. 

Other clustering approaches \cite{thrun1998clustering,liu2012transfer,carroll2005task} have also been applied to achieve transfer learning in RL. In one of the earliest notable approaches to transfer learning, Thrun et al. \cite{thrun1998clustering} described a methodology for transfer learning by clustering learning tasks using a nearest neighbor clustering approach. Although similar approaches can be used, the SOM-based approach described here preserves the topological properties of the input space, due to which similar behaviors are placed closer to one another. This could give us a rough idea of the type of behavior to be expected, given some new, arbitrary value function weights. 

Perhaps the most closely related work is the `Actor-mimic' \cite{parisotto2015actor} approach, which also performs transfer using action advice. In this approach, useful behaviors of a set of expert networks are compressed into a single multi-task network, which is then used to provide action advice in an $\epsilon -$greedy manner. The authors also report the problem of dramatically varying ranges of the value function across different tasks, which is resolved by using a Boltzmann distribution function. In the present work, the use of the cosine similarity metric resolves this issue and ensures that the similarity measure between tasks is bounded.

In the context of continual learning \cite{ring1994continual}, Ring et al. \cite{ring2011two} described a modular approach to assimilate the knowledge of complex tasks using a training process that closely resembles SOM. In this approach, a complex task is decomposed into a number of simple modules, such that modules close to each other correspond to similar agent behaviors. Teng et al. \cite{teng2015self} also proposed a SOM-based approach to integrate domain knowledge and RL, with the aim of developing agents that can continuously expand their knowledge in real time. These ideas of knowledge assimilation are also reflected in the present work. However, our approach also aims to reuse this knowledge to aid the learning of other related tasks. 


\section{Methodology}
\label{method}
In this work, we present an approach that enables the reuse of knowledge from previously learned tasks to aid the learning of a new task. 
Our approach consists of two fundamental mechanisms: (a) the accumulation of learned value function weights into a knowledge base in a scalable manner, and (b) the use of this knowledge base to guide the agent during the learning of the target task. 
The basis for these mechanisms is centered around the task similarity metric we propose here. We consider two tasks to be similar based on the cosine similarity between their corresponding learned value function weight vectors. For instance, the cosine similarity $c_{{w_{1}},{w_{2}}}$ between two non-zero weight vectors $\vec{w_{1}}$ and $\vec{w_{2}}$ is given by:
\begin{equation}
\label{eqn1}
 c_{{w_{1}},{w_{2}}}=\vec{w_{1}}.\vec{w_{2}}/|\vec{w_{1}}||\vec{w_{2}}|.
\end{equation}
The key idea is that two tasks are more likely to be similar to each other if they have similar feature weightings. Using such a similarity metric has certain advantages, such as boundedness and the ability to handle weight vectors with largely different magnitudes. That is, even in the case of highly similar or dissimilar tasks, the cosine similarity remains in the range [-1,1]. 
During the construction of the scalable knowledge base, the mentioned similarity metric is used as a basis for training the self-organizing map. 
Once this map is constructed, the cosine similarity is again used as a basis for selecting an appropriate source task weight vector to guide the exploratory behavior of the agent. We now describe these mechanisms in detail.

\subsection{Knowledge Storage Using Self-Organizing Map}

A self-organizing map (SOM)~\cite{kohonen1998self} is a type of unsupervised neural network used to produce a low-dimensional representation of its high-dimensional training samples. Typically, a SOM is represented as a two- or three-dimensional grid of nodes. Each node of the SOM is initialized to be a randomly generated weight vector of the same dimensions as the input vector. During the SOM training, an input is presented to the network, and the node that is most similar to this input is selected to be the `winner'. The winning node is then updated towards the input vector under consideration.
Other nodes in the neighborhood are also influenced in a similar manner, but as a function of their topological distances to the winner. 
The final layout of a SOM is such that adjacent nodes have a greater degree of similarity to each other in comparison to nodes that are far apart. In this way, the SOM extracts the latent structure of the input space. 

For our purposes, the knowledge of an RL task is assumed to be contained in its parameterized representation of the value function ($Q-$function), obtained using linear function approximation \cite{sutton2011reinforcement}. A na\"ive approach to storing knowledge associated with multiple tasks is to explicitly store their value function parameters/weights. Apart from the scalability issue associated with such an approach, a high degree of redundancy in the learned knowledge may arise if several of these tasks are very similar or nearly identical to each other. A more generalized approach to knowledge storage would be to store the characteristic features of the weight vectors associated with the learned tasks. The ability of the SOM to extract these features in an unsupervised manner makes it an attractive choice for the knowledge storage mechanism proposed here.

In our approach, the inputs to the SOM are learned value function weights of previously learned tasks (input tasks).
The hypothesis is that after training, the weight vectors associated with each node in the SOM have varying degrees of similarity to the input vectors, and hence, correspond to value function weights of tasks which may be related to the input tasks to varying degrees. Hence, each node in the SOM could be assumed to contain the value function information corresponding to a source task, and the weight vector associated with an appropriately selected SOM node could serve as source value function weights which could be used to guide the exploration of the agent while learning a target task.

In a continual learning scenario, a number of tasks with largely varying degrees of similarity (as per the similarity metric defined in Equation ~\eqref{eqn1}) with each other may be encountered.
A SOM containing only a few number of nodes may not be able to represent the knowledge of these tasks to a sufficient level of accuracy. Hence, the size of the SOM may need to adapt dynamically as and when new task knowledge is learned. We address this problem by allowing the number of nodes in the SOM to change, using a mechanism similar to that used in the GSOM algorithm. For a SOM containing $N$ nodes, each node $n_{i}$ is associated with an error $e_{n_{i}}$ such that for a particular input vector $\vec{w}_{v_{j}}$, if node $n_{\mathrm{win}}$ (with a corresponding weight vector $\vec{w}_{s_{n_{\mathrm{win}}}}$) is the winner, the error $e_{n_{\mathrm{win}}}$ is updated as:
\begin{equation}
\label{eqn2}
e_{n_{\mathrm{win}}}\leftarrow e_{n_{\mathrm{win}}}+1-c_{w_{v_{j}},w_{s_{n_{\mathrm{win}}}}}.
\end{equation}
The term $(1-c_{w_{v_{j}},w_{s_{n_{\mathrm{win}}}}})$ in Equation ~\eqref{eqn2} is proportional to the Euclidean distance between the $l^{2}-$ normalized versions of input vectors $\vec{w}_{v_{j}}$ and $\vec{w}_{s_{n_{\mathrm{win}}}}$. Hence, the error update equation (Equation ~\eqref{eqn2}) is equivalent to that used in \cite{alahakoon2000dynamic}. Once all the input vectors are presented to the SOM, the total error, $E$ of the network is computed as $E=\sum\limits_{i=1}^N e_{i}$. This total error is computed for each iteration of the SOM. In subsequent iterations, if the increase in the total error exceeds a certain threshold $G_{T}$, new nodes are spawned at the boundaries of the SOM. The weight vectors of these nodes are initialized to the mean of their neighbors, and are subsequently modified by the SOM training process. Such a mechanism enables the SOM to grow in size and representation capacity, thereby allowing for a low network error to be achieved.

The nature of the described SOM algorithm is such that all the input vectors are needed during the training. However, for applications such as robotics, where the agent may have limited on-board memory, this may not be feasible. Thousands of tasks may be encountered during its lifetime, and the value function weights of all these tasks would need to be explicitly stored in order to train the SOM. 
Ideally, we would like the knowledge contained in the SOM to adapt in an online manner, to include relevant information from new tasks as and when they are learned. We achieve this online adaptation by making modifications to the training mechanism of the GSOM algorithm. Specifically, when a new task is learned, we update the SOM by presenting the newly learned weights, together with the weight vectors associated with the nodes of the SOM as inputs to the GSOM algorithm. The resulting SOM is then utilized for transfer. In summary, the weights of the SOM are recycled as inputs while updating the knowledge base using the GSOM algorithm. This can be observed in the overall structure of the proposed approach, shown in Figure \ref{gen_struct}. The implicit assumption here is that the weights associated with the SOM nodes sufficiently represent the knowledge of the previously learned tasks. This approach of updating the SOM knowledge base allows new knowledge to be adaptively incorporated into the SOM, while obviating the need to explicitly store the value function weights of all previously learned tasks. The overall storage mechanism is summarized in Algorithm~\ref{alg:algorithm1}.

\begin{algorithm}[tph]
  \caption{Knowledge storage using self-organizing maps}
  \begin{algorithmic}[1]
    \STATE \textbf{Inputs}: \STATEx $\mathbf{w_{v}}=\{\vec{w}_{v_{1}}...\vec{w}_{v_{i}}...\vec{w}_{v_{M}}\}:$~A set of value function weight vectors corresponding to $M$ learned tasks. These are the input vectors to the GSOM algorithm. \STATEx $N:$~Initial number of nodes in the SOM
\STATEx $\sigma_{0}:$ Initial value of neighborhood function $\sigma$
\STATEx $\tau_{1}:$ Time constant to control the neighborhood function
\STATEx $\kappa_{0}:$ Initial value of SOM learning rate $\kappa$
\STATEx $\tau_{2}:$ Time constant to control the learning rate
\STATEx $\mathbf{w_{s}}=\{\vec{w}_{s_{1}}...\vec{w}_{s_{i}}...\vec{w}_{s_{N}}\}:$~Initial weight vectors associated with the $N$ nodes in the SOM
    \STATEx $e:$ Error vector, initialized to be zero vector of length $N$
        \STATEx $E=0:$ Initial  value of average error
                \STATEx $G_{T}:$ Growth threshold parameter
                \STATEx $N_{\mathrm{iter}}:$ Number of SOM iterations             
    \FOR {$i=1:N_{\mathrm{iter}}$}
    \STATE Randomly pick an input vector $\vec{x}$ from $\mathbf{w_{v}}$ 
    \STATE Select winning node $n_{\mathrm{win}}$ based on highest cosine similarity to input vector $x$
    \STATE $\sigma=\sigma_{0}\exp(-i/\tau_{1})$
    \STATE $\kappa=\kappa_{0}\exp(-i/\tau_{2})$
    \FOR {$j=1:N$}
    \STATE Compute topological distance $d_{n_{\mathrm{win}},j}$ between nodes $n_{\mathrm{win}}$ and $j$ 
    \STATE $h(n_{\mathrm{win}},j)=\exp(-d_{n_{\mathrm{win}},j}/2\sigma^{2})$
    \STATE $\vec{w}_{s_{j}}=\vec{w}_{s_{j}}+\kappa *h(n_{\mathrm{win}},j)*\|\vec{x}-\vec{w}_{s_{n_{\mathrm{win}}}}\|$
    \ENDFOR
    \STATE $e(n_{\mathrm{win}})=e(n_{\mathrm{win}})+1-c_{x,w_{s_{n_{\mathrm{win}}}}}$
    \STATE $E_{i}=\sum_{k=1}^{N} e_{k}$
    \IF {$(E_{i}-E_{i-1})/N>G_{T}$}
    \STATE Spawn new nodes at the boundaries of the SOM 
    \STATE Expand the error vector, with the values of new nodes initialized to the mean of the previous error vector.
    \STATE Update $N$ as per the number of new nodes added
    \ENDIF
        \ENDFOR
  \end{algorithmic}
  \label{alg:algorithm1}
\end{algorithm}




\subsection{The Transfer Mechanism}
\label{TLalgo}
Once the knowledge of previously learned tasks has been assimilated into a SOM, it is reused to aid the learning of a target task.  The weight vector associated with each SOM node is treated as the value function weight vector corresponding to an arbitrary source task. 
Among these source value function weight vectors, the one that is most similar ($w_{s_{*}}$) to the target value function weight vector $w_{T}$ is chosen for transfer. That is, $s_{*}=\underset{i\in N}{argmax} (c_{{w_{s_{i}}},{w_{T}}})$

In order to actually perform the transfer, the selected source task weights may be directly used to modify the value function weights of the target task. 
However, an insufficient degree of similarity of the source task weights could result in negative transfer. 
A safer approach is to allow the selected source value function weights to guide the exploratory actions of the agent. This guidance is provided by allowing the agent to act greedily as per the selected source value function weights with a fixed probability $\epsilon$, while exploiting the target value function that is being learned, with a probability of $1-\epsilon$. Such an approach is more unlikely to lead to drastic drops in the target task performance. 
In general, the guidance provided by the most similar source value function weights can be expected to allow the agent to execute more useful exploratory actions, thereby leading to accelerated learning of the target task.

\section{Results}

\begin{figure}[ht]
\centering
\includegraphics[width=0.8\linewidth]{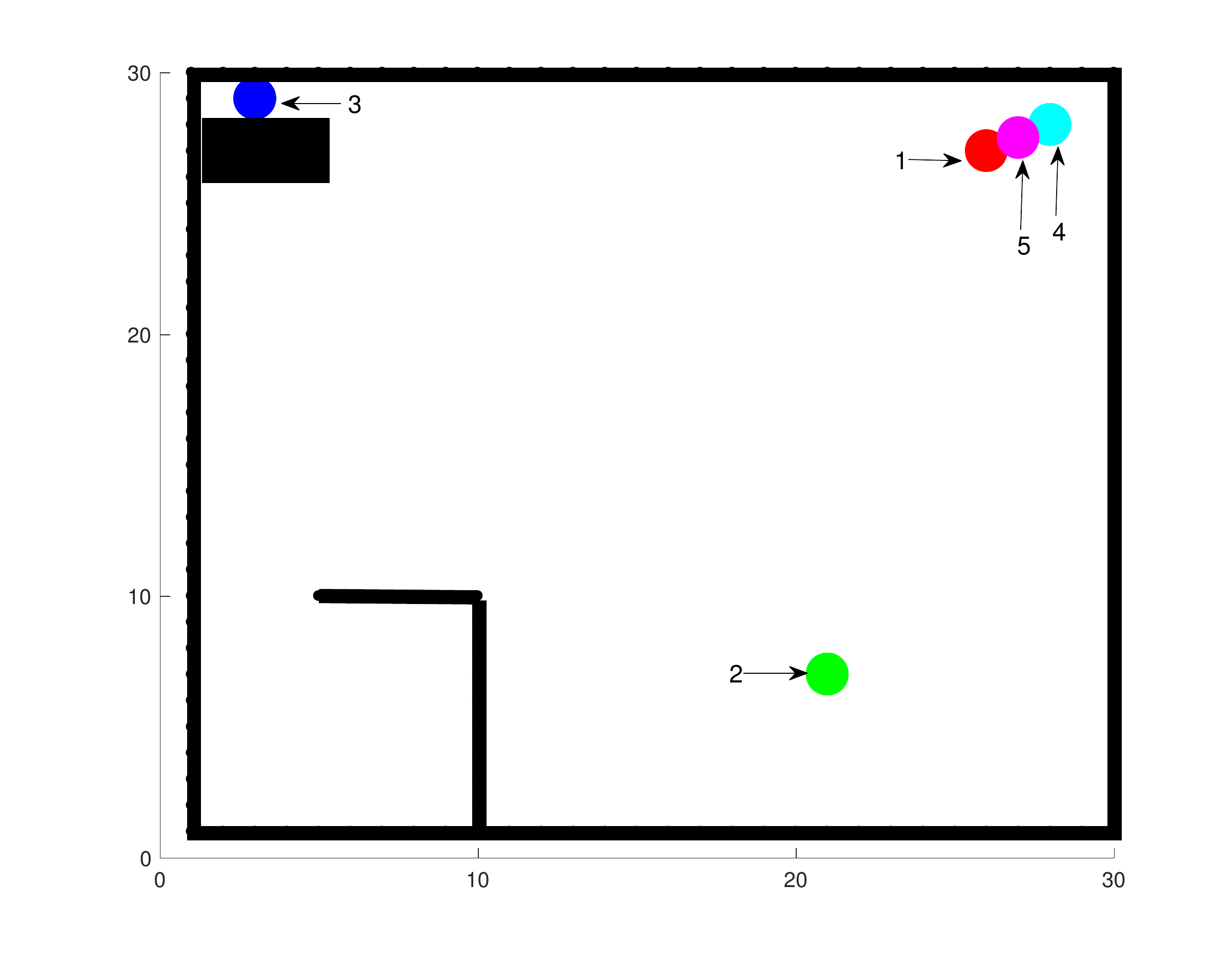}
\caption{The simulated continuous environment with the navigation goal states of different tasks (numbered from tasks $1$ to $5$), indicated by the different colored circles.   
} 
\label{env}
\end{figure}

We use the knowledge storage and reuse mechanisms described in Section \ref{method} to accelerate the learning of target tasks in a navigation environment. 
In order to evaluate the described knowledge storage and reuse mechanisms, we allow the agent to explore and learn multiple tasks in the simulated environment shown in Figure~\ref{env}. The environment is continuous, and the agent is assumed to be able to sense its horizontal and vertical coordinates, which constitute its state. The states are represented in the form of a feature vector $\vec{F_{a}}$ containing $100$ elements for each state dimension. While navigating through the environment, the agent is allowed to choose from a set of $9$ different actions: moving forwards, backwards, sideways, diagonally upwards or downwards to either side, or staying in place. The velocities associated with these movements is set to be 6 units/s, and new actions are executed every 200 ms.

As the agent executes actions in its environment, it autonomously identifies tasks using an adaptive clustering approach similar to that described in Karimpanal et al. \cite{KARIMPANAL201739}. The clustering is performed on an additional feature vector $\vec{F_{e}}$ (environment feature vector) which contains elements describing the presence or absence of specific environment features. For instance, these features could represent the presence or absence of a source of light, sound or other signals from the environment that the agent is capable of sensing.
In the simulations described here, the environment feature vector $\vec{F_{e}}$ contains $4$ elements corresponding to $4$ arbitrary environment stimuli distributed at different locations in the environment. As the agent interacts with its environment, clustering is performed on $\vec{F_{e}}$ in an adaptive manner, which helps identify unique configurations of $\vec{F_{e}}$ which may be of interest to the agent. During the agent's interactions with the environment, the mean of each discovered cluster is treated as the environment feature vector associated with the goal state of a distinct navigation task.  In our simulations, the agent eventually discovers $5$ such tasks, the corresponding goal locations of which are indicated by the colored regions in Figure ~\ref{env}. The value function corresponding to each of these tasks is learned using the $Q-\lambda$ algorithm \cite{sutton2011reinforcement}. For $Q-$learning, the reward structure is such that the agent obtains a reward ($100$) when it is in the goal state, a penalty ($-100$) for bumping into an obstacle, and a living penalty ($-10$) for every other non-goal state. In each episode, the agent starts from a random state and executes actions in the environment till it reaches the associated navigation target region (goal state), at which point, a positive reward is obtained, and the episode terminates. For each $Q-$learning task, the full feature vector $\vec{F}$ (where $\vec{F}=\{\vec{F_{e}}\cup \vec{F_{a}}\}$) is used, and the learning rate $\alpha$ is set to be $0.3$, the discount factor $\gamma$ is $0.9$ and the trace decay parameter $\lambda$ is set to be $0.9$. The other hyperparameters described in Algorithm \ref{alg:algorithm1} are set to the following values: $N=4$, $\sigma_{0}=50$, $\tau_{1}=250$, $\tau_{2}=0.1$, $G_{T}=0.3$ and $N_{iter}=1000$.

Once a new navigation task $t$ is identified, and its value function weight vector $w_{t}$ is learned, we incorporate this new knowledge into the SOM knowledge base. To do this, the value function weight vector associated with this task, along with the weight vectors associated with the SOM are presented as input vectors to Algorithm~\ref{alg:algorithm1}. For instance, if the weight vectors of the SOM are given by $\mathbf{w_{s}}=\{\vec{w}_{s_{1}}...\vec{w}_{s_{i}}...\vec{w}_{s_{N}}\}$, then the subsequent input vectors $\mathbf{w_{\mathrm{inputs}}}$ to Algorithm~\ref{alg:algorithm1} are $\mathbf{w_{\mathrm{inputs}}}=\{\mathbf{w_{s}}\cup \vec{w}_{t}\}$. By presenting the inputs to the GSOM algorithm in this manner, the resulting SOM approximates and integrates previously learned task knowledge and the knowledge of newly learned tasks. 

\begin{figure}[ht]
\centering
\includegraphics[width=1\linewidth]{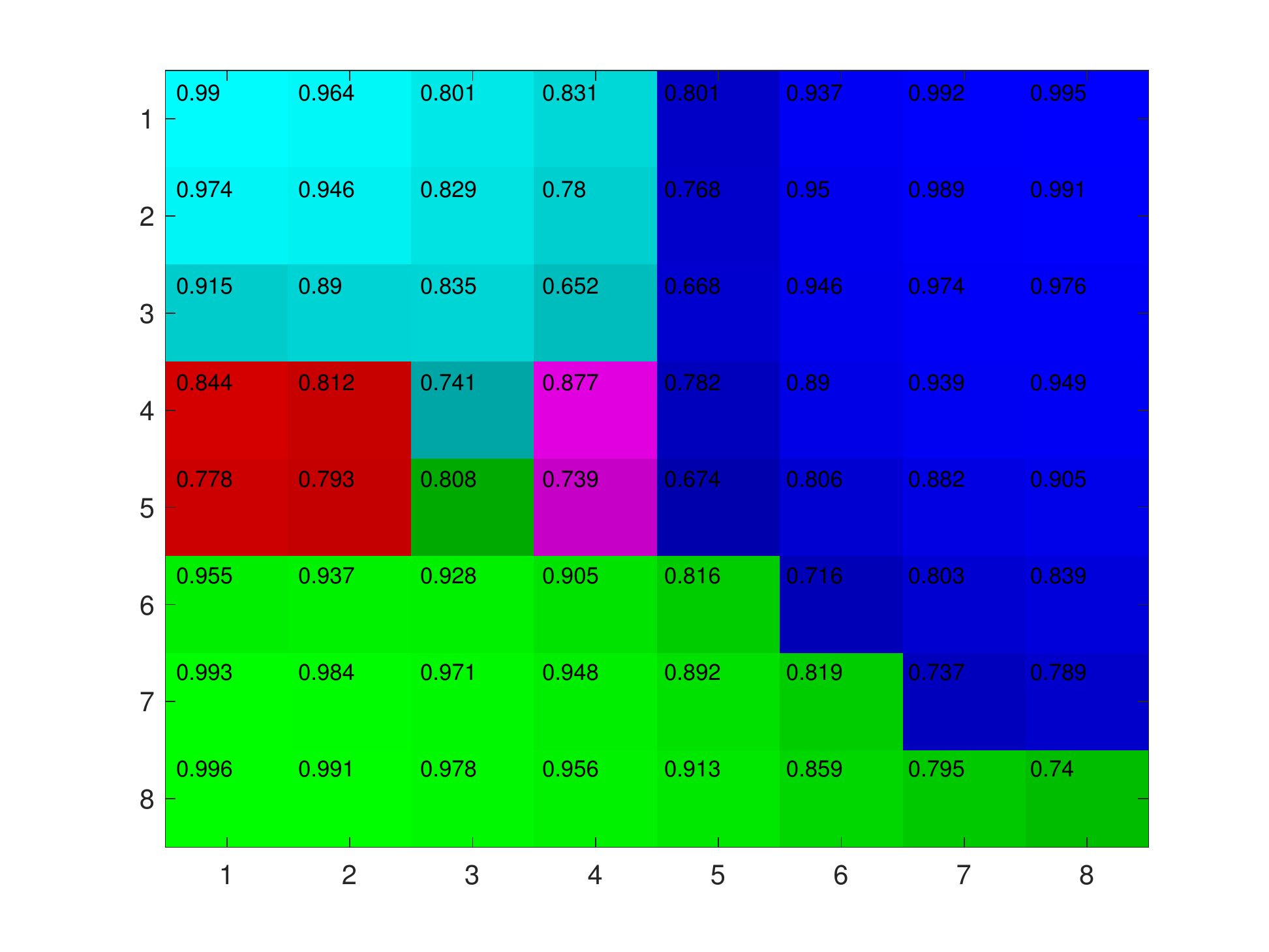}
\caption{A visual depiction of an $8\times 8$ SOM resulting from the simulations. The color of each node is derived from the most similar task in Figure \ref{env}. The intensity of the color is in proportion to the value of this similarity metric (indicated over each SOM node).
} 
\label{fig:SOMmap}
\end{figure}

Figure~\ref{fig:SOMmap} shows a sample $8\times 8$ SOM, which was learned by the agent after $1000$  $Q-$learning episodes. 
The color of each SOM node in Figure~\ref{fig:SOMmap} corresponds to the task in Figure~\ref{env} that has the maximum cosine similarity between its value function weights and the weight vector associated with the SOM node. Further, the brightness of this color is in proportion to the value of this cosine similarity. In Figure~\ref{fig:SOMmap}, these values are overlaid and displayed on top of each node. The different colors and associated cosine similarity values of each SOM node in Figure~\ref{fig:SOMmap} suggests that the SOM stores knowledge of a variety of related tasks in a structured manner.

It is also seen from Figure~\ref{fig:SOMmap} that the nodes corresponding to the knowledge of tasks that are most closely related to tasks $1,4$ and $5$ are clustered together, and those related to tasks $2$ and $3$ are distinct, and are stored in separate clusters. In addition, the allocation of the SOM nodes occurs as per the differences in the tasks themselves. For example, in the SOM shown in Figure \ref{fig:SOMmap}, $25$ nodes are allocated to tasks that are most similar to task $3$, $19$ nodes to tasks related to task $2$, and $20$ nodes to tasks related to tasks $1,4$ and $5$ combined. This demonstrates that the allocation of nodes is done as per the characteristics of the tasks, and not merely according to the number of tasks. When a number of similar tasks are learned, simply storing their value function weights would result in significant redundancies.
Such redundancies are avoided by the SOM-based approach described here.

\begin{figure}[ht]
\centering
\includegraphics[width=0.8\linewidth]{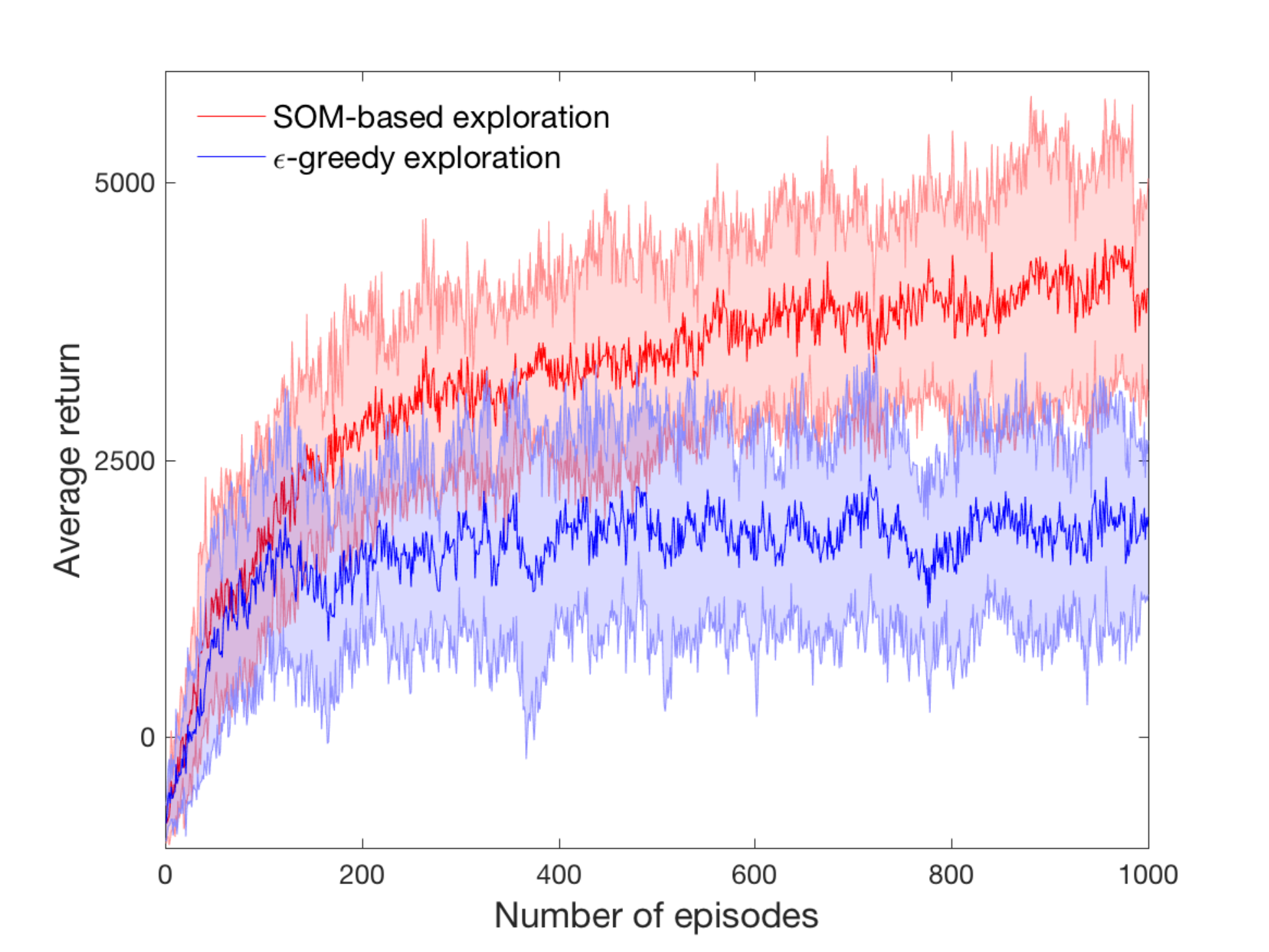}
\caption{A sample plot of the nature of the learning improvements brought about by SOM-based exploration ($\epsilon=0.3$, $G_{T}=0.3$). The solid lines represent the mean of the average return for $10$ $Q-$learning runs of $1000$ episodes each, whereas the shaded region marks the standard deviation associated with this data.
} 
\label{fig:avgreturns}
\end{figure}

Although the SOM does not necessarily retain the exact value functions of previously learned tasks, it can be used to guide the exploration of an agent while learning a new task. This is especially true if the new task is closely related to one of the previously learned tasks. Figure~\ref{fig:avgreturns} depicts this phenomenon for task $5$ ($\epsilon=0.3$, $G_{T}=0.3$), with higher returns being achieved at a significantly faster rate using the SOM-based exploration strategy in Section \ref{TLalgo}. In both exploration strategies, exploratory actions are executed with the same probability, but SOM-based exploration achieves a better performance, as knowledge of related tasks (in this case, tasks $1$ and $4$) from previous experiences allows the agent to take more informed exploratory actions.  
\begin{figure}[ht]
\centering
\includegraphics[width=0.8\linewidth]{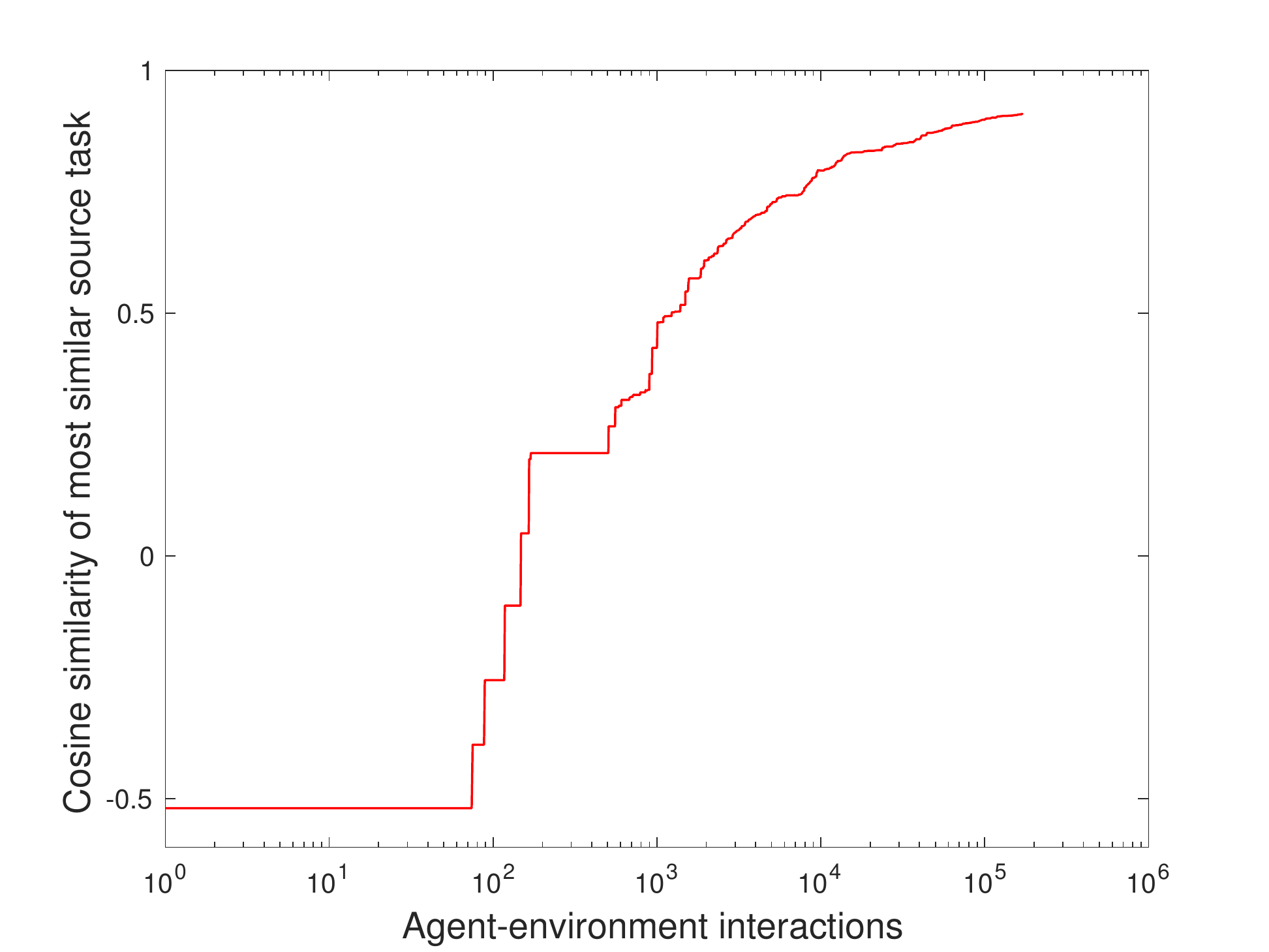}
\caption{Cosine similarity between a target task and its most similar source task as the agent interacts with its environment
} 
\label{fig:SOMadvice}
\end{figure}
This is also supported by Figure \ref{fig:SOMadvice}, which shows the evolution of the cosine similarity between the value function weights of the target task and the most similar weight vector in the SOM as the agent interacts with its environment. With a greater number of agent-environment interactions, the estimates of the agent's target task weight vector improves, and it receives more relevant advice from the SOM. This trend is probably responsible for the learning improvements seen in Figure \ref{fig:avgreturns}.

\begin{figure}[ht]
\centering
\includegraphics[width=0.9\linewidth]{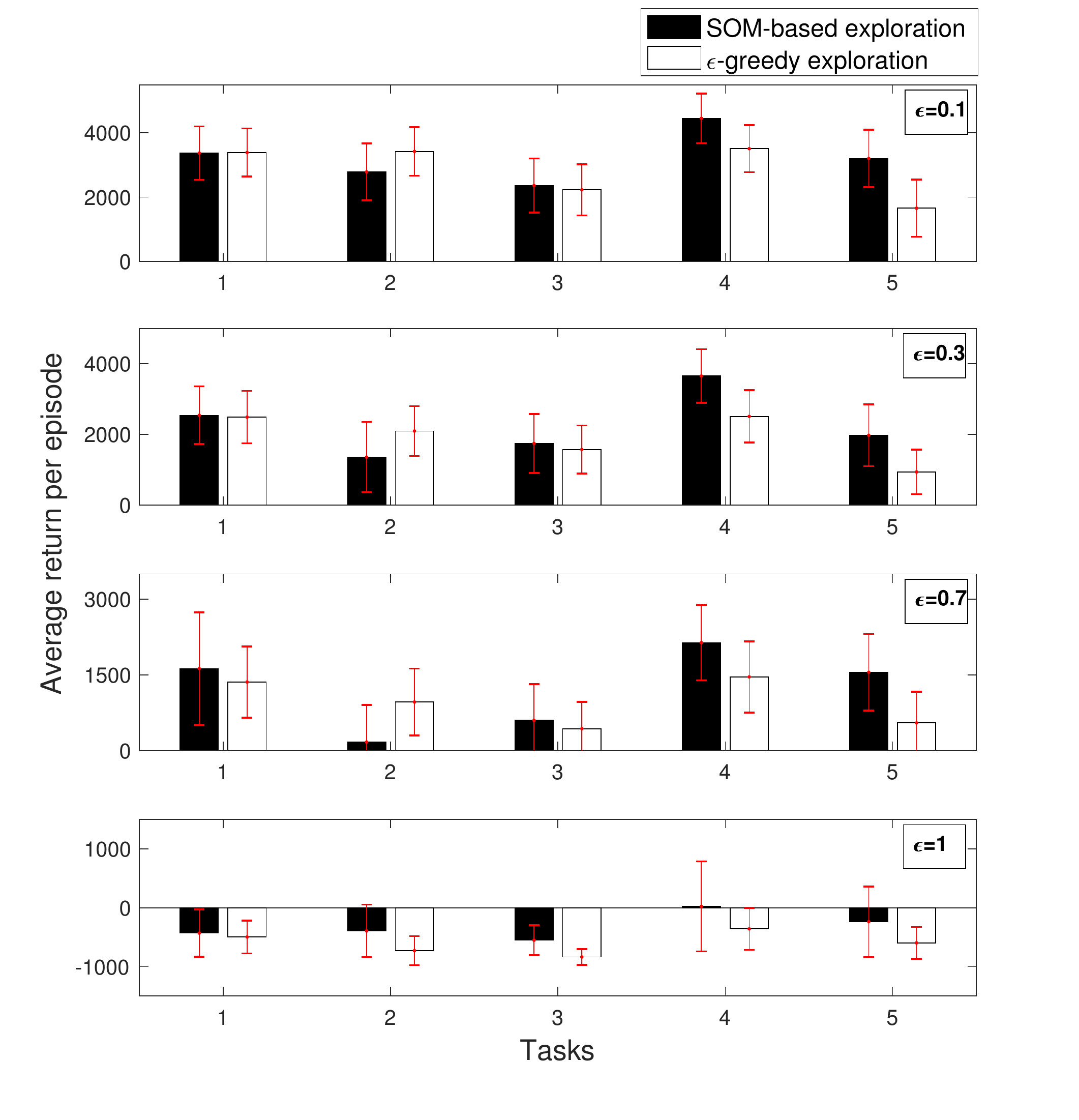}
\caption{Comparison of the average returns accumulated for different tasks using the SOM-based and $\epsilon-$ greedy exploration strategies~} 
\label{fig:returnplot}
\end{figure}

Figure~\ref{fig:returnplot} shows the average return per episode for different tasks and different values of $\epsilon$, using the two exploration strategies. The values plotted are averaged over $10$ runs. The return is computed after each episode by allowing the agent to greedily exploit the value functions starting from $100$ randomly chosen points in the environment for $100$ steps. As observed in Figure \ref{fig:returnplot}, SOM-based exploration consistently results in higher average returns for related tasks $4$ and $5$. Its performance on the unrelated tasks $2$ and $3$ are generally comparable to that of the $\epsilon-$greedy approach. Although task $1$ is related to tasks $4$ and $5$, it is the first task learned, and hence, does not benefit from the use of previous knowledge. Hence, the transfer advantage is not observed for task $1$.

\begin{figure}[ht]
\centering
\includegraphics[width=0.9\linewidth]{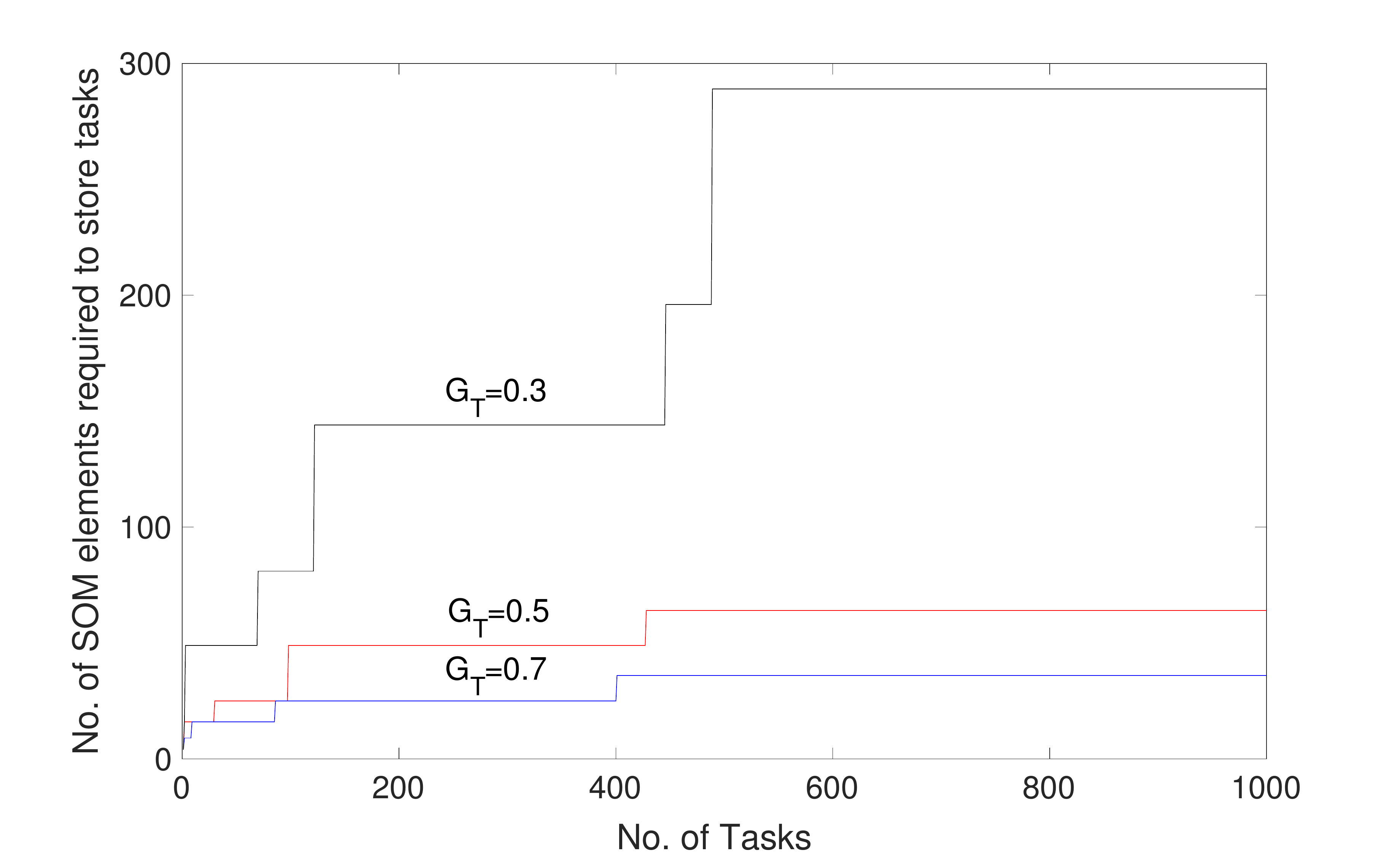}
\caption{The number of SOM nodes used to store knowledge for up to $1000$ tasks, for different values of growth threshold $G_{T}$
} 
\label{fig:SOMscaling}
\end{figure}

In addition to the improvements described, the SOM-based approach to conducting knowledge transfer also offers advantages in terms of the scalability of knowledge storage. This is depicted in Figure \ref{fig:SOMscaling}, which shows the number of nodes needed for storing the knowledge of up to $1000$ tasks, with different values of the GSOM threshold parameter $G_{T}$. It is clear that as the number of learned tasks increases, the number of nodes required per task decreases, making the SOM-based knowledge storage approach more viable. 

The simulations demonstrate that using a SOM knowledge base to guide the agent's exploratory actions help achieve faster learning when the target tasks are related to the previously learned tasks. Moreover, the nature of the transfer algorithm is such that even in the case where the source tasks are unrelated to the target task, the learning performance does not exhibit drastic drops, unlike the case where value functions of source tasks are directly used to initialize or modify the value function of a target task. Another advantage of the approach proposed here is that it can be easily applied to different representation schemes (tabular representations, neural networks etc.,), as long as the same action space and representation is used for the target and source tasks. 
In addition, with regards to the storage of knowledge of learned tasks, we demonstrated that the SOM-based approach offers a scalable alternative to explicitly storing the value function weights of all the learned tasks. 

Despite these advantages, several issues remain to be addressed. The most fundamental limitation of this approach is that it is applicable only to situations where tasks differ solely in their reward functions. This may prohibit its use in many practical applications. Moreover, the approach as described executes any action advice that it is provided with. The decision to execute the advised actions could be carried out in a more selective manner, perhaps based on the cosine similarity between the target task and the advising node of the SOM. Apart from this, and the several other possible variants to this approach, ways to automate the selection of the threshold parameters, establishing theoretical bounds on the learning performance and approaches to quantify the efficiency of the knowledge storage mechanism may be future directions for research. 

\section{Conclusions}

We described an approach to efficiently store and reuse the knowledge of learned tasks using self organizing maps. We applied this approach to an agent in a simulated multi-task navigation environment, and compared its performance to that of an $\epsilon-$greedy approach for different values of the exploration parameter $\epsilon$. Results from the simulations reveal that a modified exploration strategy that exploits the knowledge of previously learned tasks improves the agent's learning performance on related target tasks. Overall, our results indicate that the approach proposed here transfers knowledge across tasks relatively safely, while simultaneously storing relevant task knowledge in a scalable manner.
Such an approach could prove to be useful for agents that operate using the reinforcement learning framework, especially for real-world applications such as autonomous robots, where scalable knowledge storage and sample efficiency are critical factors.

\section*{Acknowledgements}
This work is partially supported by a President's Graduate Fellowship (T.G.K., Ministry of Education, Singapore) 

\bibliographystyle{ACM-Reference-Format}  
\bibliography{sample-bibliography}  


\begin{thebibliography}{00}


\ifx \showCODEN    \undefined \def \showCODEN     #1{\unskip}     \fi
\ifx \showDOI      \undefined \def \showDOI       #1{#1}\fi
\ifx \showISBNx    \undefined \def \showISBNx     #1{\unskip}     \fi
\ifx \showISBNxiii \undefined \def \showISBNxiii  #1{\unskip}     \fi
\ifx \showISSN     \undefined \def \showISSN      #1{\unskip}     \fi
\ifx \showLCCN     \undefined \def \showLCCN      #1{\unskip}     \fi
\ifx \shownote     \undefined \def \shownote      #1{#1}          \fi
\ifx \showarticletitle \undefined \def \showarticletitle #1{#1}   \fi
\ifx \showURL      \undefined \def \showURL       {\relax}        \fi
\providecommand\bibfield[2]{#2}
\providecommand\bibinfo[2]{#2}
\providecommand\natexlab[1]{#1}
\providecommand\showeprint[2][]{arXiv:#2}

\bibitem[\protect\citeauthoryear{Alahakoon, Halgamuge, and
  Srinivasan}{Alahakoon et~al\mbox{.}}{2000}]%
        {alahakoon2000dynamic}
\bibfield{author}{\bibinfo{person}{Damminda Alahakoon},
  \bibinfo{person}{Saman~K Halgamuge}, {and} \bibinfo{person}{Bala
  Srinivasan}.} \bibinfo{year}{2000}\natexlab{}.
\newblock \showarticletitle{Dynamic self-organizing maps with controlled growth
  for knowledge discovery}.
\newblock \bibinfo{journal}{{\em IEEE Transactions on neural networks\/}}
  \bibinfo{volume}{11}, \bibinfo{number}{3} (\bibinfo{year}{2000}),
  \bibinfo{pages}{601--614}.
\newblock


\bibitem[\protect\citeauthoryear{Ammar, Eaton, Taylor, Mocanu, Driessens,
  Weiss, and Tuyls}{Ammar et~al\mbox{.}}{2014}]%
        {ammar2014automated}
\bibfield{author}{\bibinfo{person}{Haitham~Bou Ammar}, \bibinfo{person}{Eric
  Eaton}, \bibinfo{person}{Matthew~E Taylor},
  \bibinfo{person}{Decebal~Constantin Mocanu}, \bibinfo{person}{Kurt
  Driessens}, \bibinfo{person}{Gerhard Weiss}, {and} \bibinfo{person}{Karl
  Tuyls}.} \bibinfo{year}{2014}\natexlab{}.
\newblock \showarticletitle{An automated measure of mdp similarity for transfer
  in reinforcement learning}. In \bibinfo{booktitle}{{\em Workshops at the
  Twenty-Eighth AAAI Conference on Artificial Intelligence}}.
\newblock


\bibitem[\protect\citeauthoryear{Carroll and Seppi}{Carroll and Seppi}{2005}]%
        {carroll2005task}
\bibfield{author}{\bibinfo{person}{James~L Carroll} {and}
  \bibinfo{person}{Kevin Seppi}.} \bibinfo{year}{2005}\natexlab{}.
\newblock \showarticletitle{Task similarity measures for transfer in
  reinforcement learning task libraries}. In \bibinfo{booktitle}{{\em Neural
  Networks, 2005. IJCNN'05. Proceedings. 2005 IEEE International Joint
  Conference on}}, Vol.~\bibinfo{volume}{2}. IEEE, \bibinfo{pages}{803--808}.
\newblock


\bibitem[\protect\citeauthoryear{Ferns, Panangaden, and Precup}{Ferns
  et~al\mbox{.}}{2004}]%
        {ferns2004metrics}
\bibfield{author}{\bibinfo{person}{Norm Ferns}, \bibinfo{person}{Prakash
  Panangaden}, {and} \bibinfo{person}{Doina Precup}.}
  \bibinfo{year}{2004}\natexlab{}.
\newblock \showarticletitle{Metrics for finite Markov decision processes}. In
  \bibinfo{booktitle}{{\em Proceedings of the 20th conference on Uncertainty in
  artificial intelligence}}. AUAI Press, \bibinfo{pages}{162--169}.
\newblock


\bibitem[\protect\citeauthoryear{Geist, Scherrer, et~al\mbox{.}}{Geist
  et~al\mbox{.}}{2014}]%
        {geist2014off}
\bibfield{author}{\bibinfo{person}{Matthieu Geist}, \bibinfo{person}{Bruno
  Scherrer}, {et~al\mbox{.}}} \bibinfo{year}{2014}\natexlab{}.
\newblock \showarticletitle{Off-policy learning with eligibility traces: a
  survey.}
\newblock \bibinfo{journal}{{\em Journal of Machine Learning Research\/}}
  \bibinfo{volume}{15}, \bibinfo{number}{1} (\bibinfo{year}{2014}),
  \bibinfo{pages}{289--333}.
\newblock


\bibitem[\protect\citeauthoryear{Karimpanal and Wilhelm}{Karimpanal and
  Wilhelm}{2017}]%
        {KARIMPANAL201739}
\bibfield{author}{\bibinfo{person}{Thommen~George Karimpanal} {and}
  \bibinfo{person}{Erik Wilhelm}.} \bibinfo{year}{2017}\natexlab{}.
\newblock \showarticletitle{Identification and off-policy learning of multiple
  objectives using adaptive clustering}.
\newblock \bibinfo{journal}{{\em Neurocomputing\/}}  \bibinfo{volume}{263}
  (\bibinfo{year}{2017}), \bibinfo{pages}{39 -- 47}.
\newblock
\showISSN{0925-2312}
\showDOI{%
\url{https://doi.org/10.1016/j.neucom.2017.04.074}}
\newblock
\shownote{Multiobjective Reinforcement Learning: Theory and Applications.}


\bibitem[\protect\citeauthoryear{Kohonen}{Kohonen}{1998}]%
        {kohonen1998self}
\bibfield{author}{\bibinfo{person}{Teuvo Kohonen}.}
  \bibinfo{year}{1998}\natexlab{}.
\newblock \showarticletitle{The self-organizing map}.
\newblock \bibinfo{journal}{{\em Neurocomputing\/}} \bibinfo{volume}{21},
  \bibinfo{number}{1} (\bibinfo{year}{1998}), \bibinfo{pages}{1--6}.
\newblock


\bibitem[\protect\citeauthoryear{Lazaric}{Lazaric}{2012}]%
        {Lazaric2012}
\bibfield{author}{\bibinfo{person}{Alessandro Lazaric}.}
  \bibinfo{year}{2012}\natexlab{}.
\newblock \bibinfo{booktitle}{{\em Transfer in Reinforcement Learning: A
  Framework and a Survey}}.
\newblock \bibinfo{publisher}{Springer Berlin Heidelberg},
  \bibinfo{address}{Berlin, Heidelberg}, \bibinfo{pages}{143--173}.
\newblock
\showISBNx{978-3-642-27645-3}
\showDOI{%
\url{https://doi.org/10.1007/978-3-642-27645-3_5}}


\bibitem[\protect\citeauthoryear{Liu, Chowdhary, How, and Carrin}{Liu
  et~al\mbox{.}}{2012}]%
        {liu2012transfer}
\bibfield{author}{\bibinfo{person}{Miao Liu}, \bibinfo{person}{Girish
  Chowdhary}, \bibinfo{person}{Jonathan~P How}, {and} \bibinfo{person}{L
  Carrin}.} \bibinfo{year}{2012}\natexlab{}.
\newblock \showarticletitle{Transfer learning for reinforcement learning with
  dependent Dirichlet process and Gaussian process}.
\newblock \bibinfo{journal}{{\em NIPS, Lake Tahoe, NV, December\/}}
  (\bibinfo{year}{2012}).
\newblock


\bibitem[\protect\citeauthoryear{Parisotto, Ba, and Salakhutdinov}{Parisotto
  et~al\mbox{.}}{2015}]%
        {parisotto2015actor}
\bibfield{author}{\bibinfo{person}{Emilio Parisotto},
  \bibinfo{person}{Jimmy~Lei Ba}, {and} \bibinfo{person}{Ruslan
  Salakhutdinov}.} \bibinfo{year}{2015}\natexlab{}.
\newblock \showarticletitle{Actor-mimic: Deep multitask and transfer
  reinforcement learning}.
\newblock \bibinfo{journal}{{\em arXiv preprint arXiv:1511.06342\/}}
  (\bibinfo{year}{2015}).
\newblock


\bibitem[\protect\citeauthoryear{Puterman}{Puterman}{1994}]%
        {Puterman:1994:MDP:528623}
\bibfield{author}{\bibinfo{person}{Martin~L. Puterman}.}
  \bibinfo{year}{1994}\natexlab{}.
\newblock \bibinfo{booktitle}{{\em Markov Decision Processes: Discrete
  Stochastic Dynamic Programming\/} (\bibinfo{edition}{1st} ed.)}.
\newblock \bibinfo{publisher}{John Wiley \& Sons, Inc.}, \bibinfo{address}{New
  York, NY, USA}.
\newblock
\showISBNx{0471619779}


\bibitem[\protect\citeauthoryear{Ring, Schaul, and Schmidhuber}{Ring
  et~al\mbox{.}}{2011}]%
        {ring2011two}
\bibfield{author}{\bibinfo{person}{Mark Ring}, \bibinfo{person}{Tom Schaul},
  {and} \bibinfo{person}{Juergen Schmidhuber}.}
  \bibinfo{year}{2011}\natexlab{}.
\newblock \showarticletitle{The two-dimensional organization of behavior}. In
  \bibinfo{booktitle}{{\em Development and Learning (ICDL), 2011 IEEE
  International Conference on}}, Vol.~\bibinfo{volume}{2}. IEEE,
  \bibinfo{pages}{1--8}.
\newblock


\bibitem[\protect\citeauthoryear{Ring}{Ring}{1994}]%
        {ring1994continual}
\bibfield{author}{\bibinfo{person}{Mark~Bishop Ring}.}
  \bibinfo{year}{1994}\natexlab{}.
\newblock {\em \bibinfo{title}{Continual learning in reinforcement
  environments}}.
\newblock \bibinfo{thesistype}{Ph.D. Dissertation}. \bibinfo{school}{University
  of Texas at Austin Austin, Texas 78712}.
\newblock


\bibitem[\protect\citeauthoryear{Song, Gao, Wang, and An}{Song
  et~al\mbox{.}}{2016}]%
        {song2016measuring}
\bibfield{author}{\bibinfo{person}{Jinhua Song}, \bibinfo{person}{Yang Gao},
  \bibinfo{person}{Hao Wang}, {and} \bibinfo{person}{Bo An}.}
  \bibinfo{year}{2016}\natexlab{}.
\newblock \showarticletitle{Measuring the distance between finite markov
  decision processes}. In \bibinfo{booktitle}{{\em Proceedings of the 2016
  International Conference on Autonomous Agents \& Multiagent Systems}}.
  International Foundation for Autonomous Agents and Multiagent Systems,
  \bibinfo{pages}{468--476}.
\newblock


\bibitem[\protect\citeauthoryear{Sutton and Barto}{Sutton and Barto}{2011}]%
        {sutton2011reinforcement}
\bibfield{author}{\bibinfo{person}{Richard~S Sutton} {and}
  \bibinfo{person}{Andrew~G Barto}.} \bibinfo{year}{2011}\natexlab{}.
\newblock \bibinfo{title}{Reinforcement learning: An introduction}.
\newblock   (\bibinfo{year}{2011}).
\newblock


\bibitem[\protect\citeauthoryear{Sutton, Modayil, Delp, Degris, Pilarski,
  White, and Precup}{Sutton et~al\mbox{.}}{2011}]%
        {sutton2011horde}
\bibfield{author}{\bibinfo{person}{Richard~S Sutton}, \bibinfo{person}{Joseph
  Modayil}, \bibinfo{person}{Michael Delp}, \bibinfo{person}{Thomas Degris},
  \bibinfo{person}{Patrick~M Pilarski}, \bibinfo{person}{Adam White}, {and}
  \bibinfo{person}{Doina Precup}.} \bibinfo{year}{2011}\natexlab{}.
\newblock \showarticletitle{Horde: A scalable real-time architecture for
  learning knowledge from unsupervised sensorimotor interaction}. In
  \bibinfo{booktitle}{{\em The 10th International Conference on Autonomous
  Agents and Multiagent Systems-Volume 2}}. International Foundation for
  Autonomous Agents and Multiagent Systems, \bibinfo{pages}{761--768}.
\newblock


\bibitem[\protect\citeauthoryear{Taylor and Stone}{Taylor and Stone}{2009}]%
        {taylor_transfer_2009}
\bibfield{author}{\bibinfo{person}{Matthew~E Taylor} {and}
  \bibinfo{person}{Peter Stone}.} \bibinfo{year}{2009}\natexlab{}.
\newblock \showarticletitle{Transfer learning for reinforcement learning
  domains: A survey}.
\newblock \bibinfo{journal}{{\em Journal of Machine Learning Research\/}}
  \bibinfo{volume}{10}, \bibinfo{number}{Jul} (\bibinfo{year}{2009}),
  \bibinfo{pages}{1633--1685}.
\newblock


\bibitem[\protect\citeauthoryear{Teng, Tan, and Zurada}{Teng
  et~al\mbox{.}}{2015}]%
        {teng2015self}
\bibfield{author}{\bibinfo{person}{Teck-Hou Teng}, \bibinfo{person}{Ah-Hwee
  Tan}, {and} \bibinfo{person}{Jacek~M Zurada}.}
  \bibinfo{year}{2015}\natexlab{}.
\newblock \showarticletitle{Self-organizing neural networks integrating domain
  knowledge and reinforcement learning}.
\newblock \bibinfo{journal}{{\em IEEE transactions on neural networks and
  learning systems\/}} \bibinfo{volume}{26}, \bibinfo{number}{5}
  (\bibinfo{year}{2015}), \bibinfo{pages}{889--902}.
\newblock


\bibitem[\protect\citeauthoryear{Thrun and O'Sullivan}{Thrun and
  O'Sullivan}{1998}]%
        {thrun1998clustering}
\bibfield{author}{\bibinfo{person}{Sebastian Thrun} {and}
  \bibinfo{person}{Joseph O'Sullivan}.} \bibinfo{year}{1998}\natexlab{}.
\newblock \showarticletitle{Clustering learning tasks and the selective
  cross-task transfer of knowledge}.
\newblock In \bibinfo{booktitle}{{\em Learning to learn}}.
  \bibinfo{publisher}{Springer}, \bibinfo{pages}{235--257}.
\newblock


\bibitem[\protect\citeauthoryear{Torrey and Taylor}{Torrey and Taylor}{2013}]%
        {torrey2013teaching}
\bibfield{author}{\bibinfo{person}{Lisa Torrey} {and} \bibinfo{person}{Matthew
  Taylor}.} \bibinfo{year}{2013}\natexlab{}.
\newblock \showarticletitle{Teaching on a budget: Agents advising agents in
  reinforcement learning}. In \bibinfo{booktitle}{{\em Proceedings of the 2013
  international conference on Autonomous agents and multi-agent systems}}.
  International Foundation for Autonomous Agents and Multiagent Systems,
  \bibinfo{pages}{1053--1060}.
\newblock


\bibitem[\protect\citeauthoryear{White, Modayil, and Sutton}{White
  et~al\mbox{.}}{2012}]%
        {white2012scaling}
\bibfield{author}{\bibinfo{person}{Adam White}, \bibinfo{person}{Joseph
  Modayil}, {and} \bibinfo{person}{Richard~S Sutton}.}
  \bibinfo{year}{2012}\natexlab{}.
\newblock \showarticletitle{Scaling life-long off-policy learning}. In
  \bibinfo{booktitle}{{\em Development and Learning and Epigenetic Robotics
  (ICDL), 2012 IEEE International Conference on}}. IEEE, \bibinfo{pages}{1--6}.
\newblock


\bibitem[\protect\citeauthoryear{Zhan and Taylor}{Zhan and Taylor}{2015}]%
        {zhan2015online}
\bibfield{author}{\bibinfo{person}{Yusen Zhan} {and} \bibinfo{person}{Matthew~E
  Taylor}.} \bibinfo{year}{2015}\natexlab{}.
\newblock \showarticletitle{Online transfer learning in reinforcement learning
  domains}.
\newblock \bibinfo{journal}{{\em arXiv preprint arXiv:1507.00436\/}}
  (\bibinfo{year}{2015}).
\newblock


\bibitem[\protect\citeauthoryear{Zimmer, Viappiani, and Weng}{Zimmer
  et~al\mbox{.}}{2014}]%
        {zimmer2014teacher}
\bibfield{author}{\bibinfo{person}{Matthieu Zimmer}, \bibinfo{person}{Paolo
  Viappiani}, {and} \bibinfo{person}{Paul Weng}.}
  \bibinfo{year}{2014}\natexlab{}.
\newblock \showarticletitle{Teacher-student framework: a reinforcement learning
  approach}. In \bibinfo{booktitle}{{\em AAMAS Workshop Autonomous Robots and
  Multirobot Systems}}.
\newblock


\end{thebibliography}

\end{document}